\documentclass[runningheads]{llncs}
\usepackage{amsmath} 
\usepackage{graphicx}
\usepackage{tabularx}
\usepackage{multirow}
\usepackage{url}
\usepackage{caption}
\usepackage[table]{xcolor}
\usepackage{xcolor}
\usepackage{soul}
\usepackage{courier}
\usepackage[T1]{fontenc}
\usepackage{epstopdf}
\newcolumntype{Y}{>{\centering\arraybackslash}X}
\usepackage{collcell}
\usepackage{xfp}

\usepackage{graphicx}
\usepackage{tabularx}
\makeatletter
\newcommand{\printfnsymbol}[1]{%
  \textsuperscript{\@fnsymbol{#1}}%
}
\makeatother

\begin{document}
\title{Designing a Socially Assistive Robot \\to Support Older Adults with Low Vision}
%
%
\author{Emily Zhou\thanks{equal contribution}\inst{1} \and
Zhonghao Shi\printfnsymbol{1}\inst{1}\and
Xiaoyang Qiao\inst{1} \and
Maja J Matari\'c\inst{1} \and
Ava K. Bittner\inst{2} }

\authorrunning{E. Zhou et al.}
%

\institute{University of Southern California, USA \email{\{emilyzho, zhonghas,xiaoyanq, mataric\}@usc.edu} \and
University of California, Los Angeles, USA \email{	abittner@mednet.ucla.edu} }
%




%
\maketitle              
\begin{abstract}
Socially assistive robots (SARs) have shown great promise in supplementing and augmenting interventions to support the physical and mental well-being of older adults. However, past work has not yet explored the potential of applying SAR to lower the barriers of long-term low vision rehabilitation (LVR) interventions for older adults. In this work, we present a user-informed design process to validate the motivation and identify major design principles for developing SAR for long-term LVR. To evaluate user-perceived usefulness and acceptance of SAR in this novel domain, we performed a two-phase study through user surveys.  First, a group (n=38) of older adults with LV completed a mailed-in survey.  Next, a new group (n=13) of older adults with LV saw an in-clinic SAR demo and then completed the survey. The study participants reported that SARs would be useful, trustworthy, easy to use, and enjoyable while providing socio-emotional support to augment LVR interventions. The in-clinic demo group reported significantly more positive opinions of the SAR's capabilities than did the baseline survey group that used mailed-in forms without the SAR demo. 

\keywords{socially assistive robotics  \and low vision rehabilitation.}
\end{abstract}

\begin{figure}[h!]
\centerline{\includegraphics[scale=0.25]{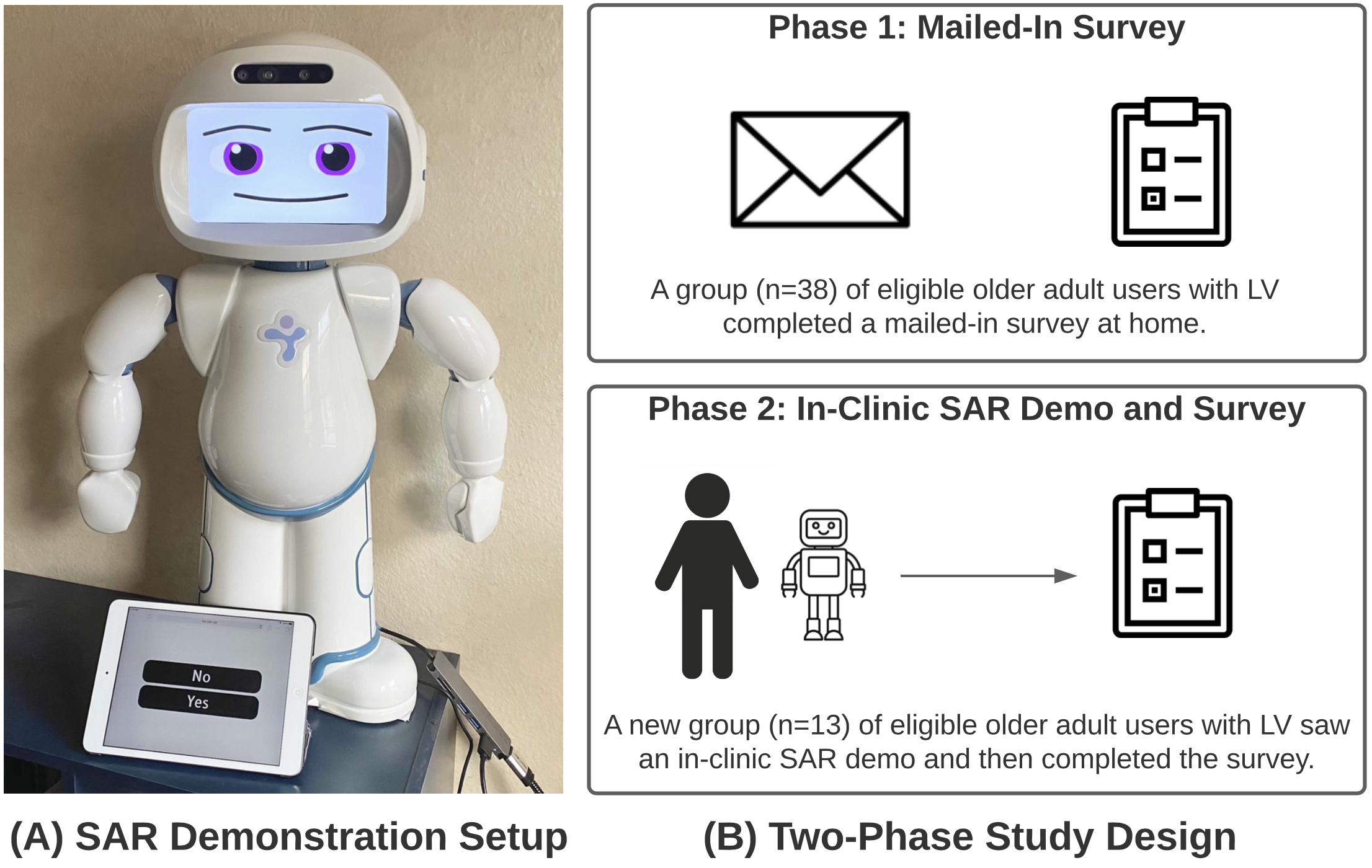}}
\caption{\textbf{(A)} Prototype SAR demonstration system setup deployed in phase 2 of the study; \textbf{(B)} Overview of the two-phase study design.}
\label{fig:study}
\end{figure}
\section{Introduction}
Past research has shown that low vision (LV) cannot be corrected with traditional treatments, but visual functioning can improve significantly with magnifier and LV rehabilitation (LVR) training at follow-up visits beyond the initial visit at which magnifiers are dispensed~\cite{stelmack2017outcomes}. The successful application of magnifiers for reading is predicated on patient adherence and motivation for correct and sustained use, which often requires LVR training following the acquisition of the magnifier. However, physical and financial barriers prevent the provision of in-clinic LVR services~\cite{pollard2003barriers}. For these reasons, many individuals with LV do not return for training to become proficient in the use of magnifiers for important tasks, such as reading~\cite{rubin2013measuring}. To tackle these challenges, this work presents a user-informed design process to validate the motivation and identify design principles for developing an SAR to foster the use of magnifiers for reading at home during daily activities by patients with LV through a long-term intervention. To evaluate user-perceived usefulness and acceptance of SAR in this novel domain, we conducted a two-phase study to collect quantitative and qualitative survey data: 1) an initial mailed-in survey without a SAR demonstration (demo); 2) an in-clinic SAR demo and survey. The two major contributions of this work are summarized below:

\textbf{1) Identified and validated the motivation for this new SAR domain:} This paper introduces a novel SAR interaction paradigm to foster long-term LVR interventions toward more frequent, longer duration, and improved use of magnifiers in the daily activities of users with LV. The results from the quantitative survey responses collected in a two-phase study provide evidence to support the development of SARs for supporting users' socio-emotional needs during LVR.

\textbf{2) Developed a set of user-informed design guidelines:} To inform future SAR development, this paper analyzes qualitative survey data collected from the in-clinic demo (Phase 2). Based on LV participants' self-reported preferences and expectations, an inductive coding process~\cite{thomas2006general} was used to generate a set of user-informed design guidelines to inform future development of SARs in this new domain.

    

\section{Background and Related Work}

We briefly overview most relevant research in LVR and SARs for older adults.

\textbf{Low Vision Rehabilitation (LVR):} LV can lead to reduced quality of life and increased depression and/or emotional distress~\cite{goldstein2012baseline}.  Fortunately, a large body of research has shown that LVR can improve socio-emotional aspects and functional ability~\cite{binns2012effective}. However, skills taught by LVR providers in-clinic may not translate to the home without persistent in-home practice and continued support~\cite{stelmack2017outcomes}. New magnifiers are abandoned within the first three months by about one in five users with LV when they are perceived as ineffective for the task~\cite{dougherty2011abandonment}, which may be preventable with additional LVR to maximize visual functioning~\cite{stelmack2017outcomes}. Although most LVR providers agree that the patient’s home would be the optimal setting for providing LVR services~\cite{liu2013occupational}, there are physcial and financial challenges related to the provision of both home visits and tele-rehabilitation~\cite{bittner2018feasibility}. Given these barriers~\cite{pollard2003barriers}, it is estimated that only 10-20\% of the population in the developed world has access to LVR~\cite{chiang2011global}. For this reason, it is imperative to validate novel solutions for providing LVR, such as SARs as a complementary approach, in order to overcome the existing barriers and challenges that limit LVR care.

 \textbf{Socially Assistive Robotics (SAR) for Older Adults:} SARs have shown great potential for providing cost-effective health and social support for older adults~\cite{abdi2018scoping}. Compared to other conventional and technology-based solutions, such as mobile applications, research has shown that the physical embodiment of SARs helps to foster social rapport and emotional engagement with users,  enabling more effective delivery of behavioral interventions~\cite{Deng2019EmbodimentIS}. In addition, studies have shown that SARs can help older adults to significantly increase exercise~\cite{gadde2011toward} and medication adherence~\cite{rantanen2017home} in long-term in-home settings, while reporting increased positive attitude toward using SAR. Despite this progress in developing SARs for older adults, the potential to apply SARs for in-home LVR interventions has not yet been explored. Our aim in this work was to introduce and validate a novel interaction paradigm to develop a SAR specifically for users who could benefit from LVR support.


\section{Experimental Design}

As shown in Fig.~\ref{fig:study}, a two-phase study with two cohorts of participants evaluated user-perceived usefulness and acceptance of a SAR for LVR, as well as  generated a set of user-informed design guidelines for this new SAR domain. The study was approved by the USC Institutional Review Board under protocol \#UP-20-00359.

 \textbf{Participants:} Subjects were recruited with the following criteria: 1) English-speaking/fluent adults with LV who were seen at the UCLA Vision Rehabilitation Center, 2) used a magnifier, and 3) had no severe hearing loss.


\textbf{Survey Methodology: } 
Based on surveys validated in past SAR research~\cite{beuscher2017socially}, a set of 7-point Likert scale questions were designed to inquire about the difficulty or frustration with magnifiers, obtain users' self-ratings of vision and general health, and assess user-perceived usefulness and acceptance of SARs for LV.
The survey questions are shown in Table~\ref{tab:questions}.
The same questions were used in both phases of the study.
The in-clinic phase (Phase 2) also  included additional semi-structured interview questions to obtain participants’ qualitative feedback about the SAR via suggestions for additional features or other content to improve the SAR’s likeability or usability.

\begin{table}[t!]
\caption{Likert scale survey questions used in both study phases.} 
\begin{tabular}{|p{3cm}|p{9cm}|}
\hline
{\textbf{Topics} }& 
\textbf{7-point Likert Scale Survey Questions} \\
\hline
{User's vision} & 
My vision is very poor. \\
\hline 
{User's perception of magnifier use} & 
It was difficult when first learning to use a new magnifier for reading. \newline
I have been frustrated when using magnifiers for reading. \\
\hline
{User-perceived usefulness of SARs for LVR}  & 
A robot would be useful when first learning to use a magnifier. \newline
It would be a good idea to use a robot to help with vision loss. \\
\hline
{User-perceived acceptance}  & 
I would trust a robot to give good advice about magnifier use. \newline
I think I would find the robot easy to use. \newline
I would enjoy the robot talking to me. \\
\hline

\end{tabular}
\label{tab:questions}
\end{table}

\textbf{Phase 1 Mailed-In Survey:}
The quantitative Likert scale survey was mailed to older adults users with LV who had recently purchased a magnifier. The survey provided an image of a SAR and a general description; the participants did not have the opportunity for an in-clinic/in-person demonstration of the SAR. A total of 38 participants took part in Phase 1 by returning anonymous survey responses by mail. 

\textbf{Phase 2 In-Clinic SAR Demo and Survey: } 
In Phase 2, a new cohort of participants was recruited for an in-clinic visit and  demo interaction with a SAR. After the demo, they completed a survey that included both quantitative Likert scale questions from Phase 1 and new qualitative semi-structured interview questions. A total of 13 new participants took part in Phase 2. The SAR demo interaction consisted of SAR’s self-introduction, followed by initial questions about the participant’s vision and magnifier use. The SAR asked the participant questions such as "How do you feel about your vision right now?" Participants were prompted to use an iPad Mini tablet interface to enter their multipe-choice responses via large buttons with reversed contrast and enlarged text for LV. Based on the participant's answers, the SAR responded with praise, sympathy, and/or encouragement to provide the appropriate socio-emotional support. We also incorporated entertaining dialogue in the form of jokes and a short, positive news story, aiming to increase the SAR's likeability.

\textbf{Study Hypotheses:} Based on the results of previous research on SARs for older adults and the study team's clinical expertise, we developed the following hypotheses related to the need for and usability of a SAR to facilitate LVR:
\begin{description}
    \item[$\bullet$ H1 (Experiences with Magnifiers):] We hypothesized that the majority of participants would indicate they had difficulty \textbf{(H1-a)} and frustration \textbf{(H1-b)} when first learning to use a magnifier for reading. We also anticipated that participants with self-perceived very poor vision would be more likely to report difficulty \textbf{(H1-c)} and frustration \textbf{(H1-d)} with magnifier use.

\item[$\bullet$ H2 (Perceived Acceptance of SAR for LVR):] Without an in-clinic interaction with the SAR, we hypothesized that Phase 1 participants with LV in the mailed-in group would still report that a SAR could be useful for magnifier learning~\textbf{(H2-a)} and facilitating LVR~\textbf{(H2-b)}. In addition, we expected that participants who self-reported very poor vision would be more likely to think a SAR would be useful for facilitating magnifier use~\textbf{(H2-c)} and LVR~\textbf{(H2-d)}. We also anticipated that LV participants would perceive the SAR robot to be trustworthy~\textbf{(H2-e)}, easy to use~\textbf{(H2-f)}, and enjoyable~\textbf{(H2-g)}.

\item[$\bullet$ H3 (Mailed-In vs. In-Clinic Post-Demo Survey Responses):] We hypothesized that having an in-clinic interaction with the SAR would help Phase 2 participants develop similar or more positive opinions than Phase 1 participants for the following topics: the usefulness of a SAR for first-time magnifier use~\textbf{(H3-a)}, help with vision loss~\textbf{(H3-b)},  trust in the SAR's advice about magnifiers~\textbf{(H3-c)}, ease of use of the SAR~\textbf{(H3-d)}, and enjoyability of the SAR~\textbf{(H3-e)}.
\end{description}

\section{Methods}

\subsection{SAR System Implementation}
\begin{figure} [t!]
\centerline{\includegraphics[scale=.28]{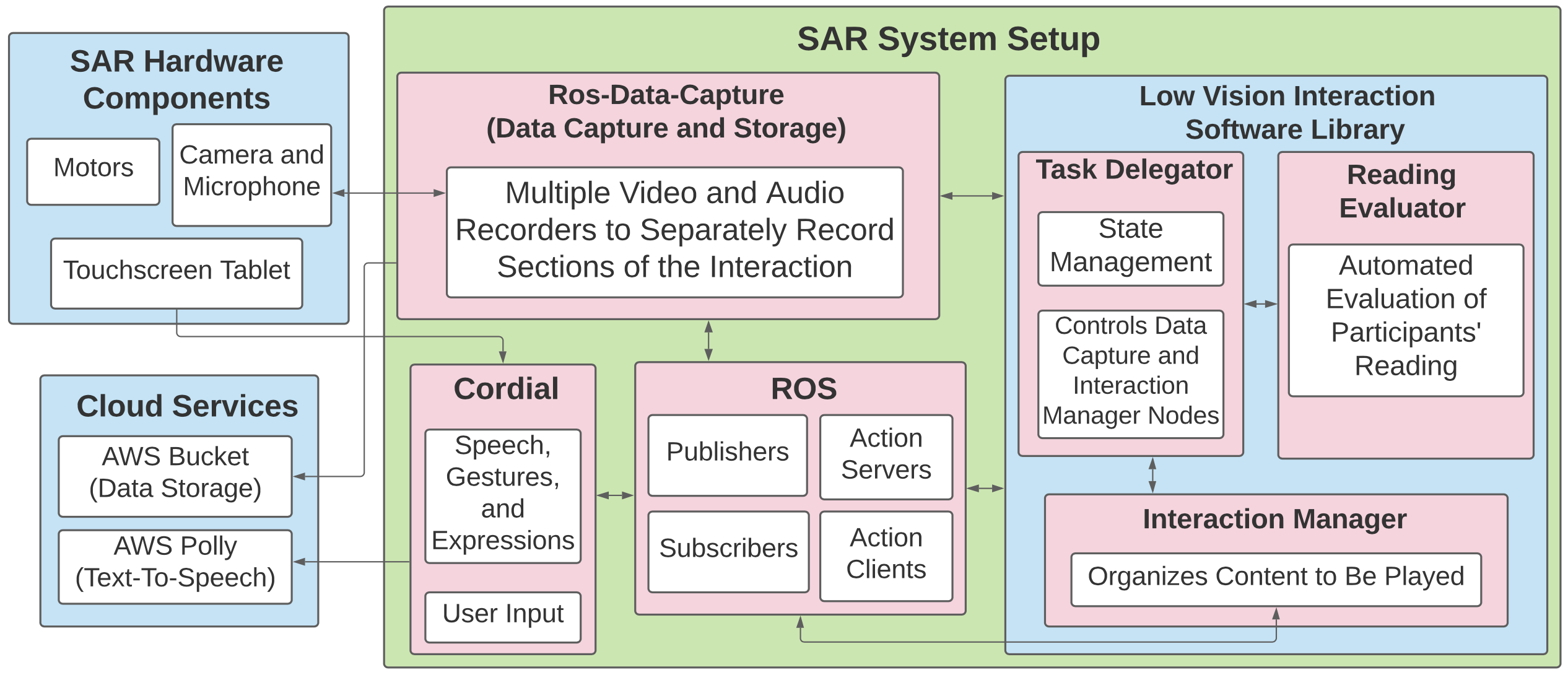}}
\caption{System architecture of our SAR system hardware and software.}
\label{fig:system}
\end{figure}

As shown in Fig.~\ref{fig:system}, the developed prototype SAR system consisted of LuxAI’s non-mobile humanoid QTrobot (QT)~\cite{qtrobot} and a tablet with a graphical user interface (GUI) in a large, high-contrast Sans Serif font to enable easy readability for user with LV. We developed a ROS-based~\cite{quigley2009ros} software system in Python, available at \url{https://github.com/robotpt/vision-project.git}, to manage the social human-robot interaction. As shown in Fig.~\ref{fig:system}, we also leveraged existing libraries: 1) Cordial~\cite{short2020sprite} to coordinate robot speech, gestures, and facial expressions; 2) ros-data-capture with Amazon Web Services (AWS)~\cite{mathew2014overview} and MongoDB~\cite{chodorow2013mongodb} to handle data collection and storage; and 3) AWS Polly neural text-to-speech service~\cite{mathew2014overview} to synthesize the robot's dialog.

\subsection{Survey Analysis}
This subsection provides an overview of the methods used for both the quantitative and qualitative survey data analyses.

\textbf{Quantitative analyses (Statistical Tests):} The following statistical tests were performed to evaluate the research hypotheses about user's past experience with magnifier use and their acceptance of using the SAR to facilitate LVR interventions: 1) one-sample Wilcoxon signed rank test (\textbf{H1,H2}), with the null hypothesis being defined as when the median was greater than or equal to 4 (neutral) on the 7-point Likert scale; 2) ordinal logistic regression (\textbf{H1,H2}); and 3) the two-tailed Mann-Whitney U test (\textbf{H3}). More details of the descriptive statistical analysis are reported in~Sec.~\ref{sec:results}.

\textbf{Qualitative Analyses (Inductive Coding Process):} In order to identify the major user-informed design principles for future SAR development, we also followed the process of inductive coding~\cite{thomas2006general} to analyze the qualitative survey responses collected from the semi-structured interview questions administered in Phase 2. All the responses were transcribed, summarized, labeled, and finalized with the corresponding design principles based on participants' suggestions.

\section{Results}
\label{sec:results}

\begin{figure}[t!]
\centerline{\includegraphics[scale=0.4]{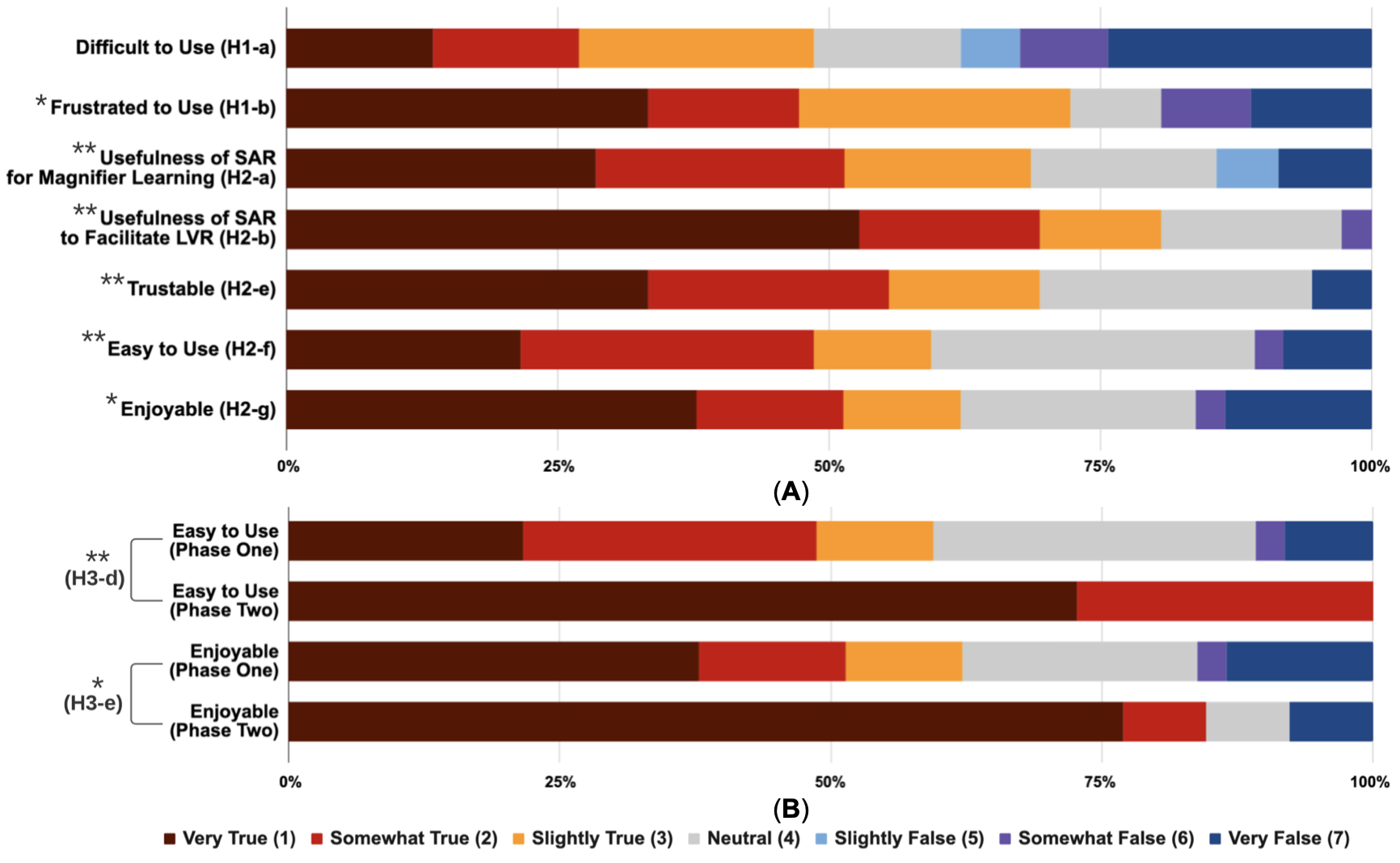}}
\caption{ \textbf{(A)} A stacked bar graph of participants' ratings relevant to H1 and H2. The majority of participants felt frustrated (72.2\%) about magnifier use, and felt that a SAR would be useful (68.6 \%), a good idea (80.6 \%), and trustworthy (69.4 \%) to facilitate LVR, especially for their socio-emotional needs. \textbf{(B)} A comparison of stacked bar graphs between the Phase 1 mailed-in and Phase 2 in-clinic demo groups, showing that the Phase 2 group developed more positive opinions of SAR's capabilities than the Phase 1 group (* = \textit{p} < 0.05, ** = \textit{p} < 0.001).}
\label{fig:stacked}
\end{figure}

As shown in Fig.~\ref{fig:stacked} and Table~\ref{tab:design}, this section reports the findings from both the quantitative and qualitative survey analyses. Participants were older adult users with LV with an average age of 74 (SD = 17).

\textbf{H1 (Experiences with Magnifiers):} As determined by the one-sample Wilcoxon signed rank test, the responses for the magnifier being difficult to use (\textbf{H1a}, median = 4, \textit{p} = .223) were not found to be significantly below 4, but the responses for feeling frustrated with magnifier use (\textbf{H1b}, median = 2, \textit{p} = .015) were found to be significantly below 4. Ordinal logistic regressions were used to evaluate the relationships between self-perceived vision level (independent variable) and the perception of magnifiers being difficult (\textbf{H1-c}, \textit{p} = .177) or frustrating to use (\textbf{H1-d}, \textit{p} = .465) (dependent variables). 



\textbf{H2 (Perceived Acceptance of SARs for LVR):} To evaluate the participants' acceptance of a SAR for LVR, the responses for perceiving it to be useful for learning to use a magnifier (\textbf{H2-a}, median = 2, \textit{p} = .002) and a good idea to help with LV (\textbf{H2-b}, median = 1, \textit{p} < .001) were found to be significantly below 4 as determined by the one-sample Wilcoxon signed rank test. Based on ordinal logistic regression analyses, we found that self-perceived vision level was not significantly associated with the user-perceived usefulness of a SAR to support new magnifier use (\textbf{H2-c}, z = 1.215, \textit{p} = .224), but self-reported vision level was significantly related to the perception that the SAR was a good idea to help with vision loss (\textbf{H2-d}, z = 3.095, \textit{p} = .002), as those who had worse vision were more likely to believe it was a good idea. In addition, we found that the responses for trusting the SAR's advice about magnifiers (\textbf{H2-e}, median = 2, \textit{p} < .001), the SAR's ease of use (\textbf{H2-f}, median = 3, \textit{p} = .008), and enjoyment of the SAR's talking (\textbf{H2-g}, median = 2, \textit{p} = .014) were all significantly below 4 as determined by the one-sample Wilcoxon signed rank test, further validating that LV participants had a high level of acceptance of a SAR for LVR.


\textbf{H3 (Mailed-in vs. In-Clinic Post-Demo Survey Responses):} To assess whether participants' perceptions of the SAR were different with versus without a demo interaction, two-tailed Mann-Whitney U tests were used to compare the responses from the mailed-in and in-clinic demo groups. We found no significant differences in responses between the two groups in terms of perceived usefulness for new magnifier support (\textbf{H3-a}, U = 199, \textit{p} = .783), help with LV (\textbf{H3-b}, U = 169.5, \textit{p} = .215), or the perceived trust in the SAR to give good advice about magnifier use (\textbf{H3-c}, U = 214, \textit{p} = .638). However, we did find significant differences for whether the SAR would be easy between the the mailed-in (Median = 3) and in-clinic post-demo (Median = 1) surveys (\textbf{H3-d}, U = 71, \textit{p} < .001), as well as for enjoyability of the SAR talking between the mailed-in (Median = 2) versus in-clinic post-demo (Median = 1) surveys (\textbf{H3-e}, U = 148, \textit{p} = .029).

\begin{table*}[t!]
\caption{Design principles identified as relevant for the development of SAR to support older adults with low vision} 
\centering 
\begin{tabularx}{\textwidth}{|X|X|}
\hline
 \textbf{Examples of Participants' Quotes} & \textbf{Design Principles} \\
\hline
"Give it a clearer/firmer voice"; "consider changing voice, QT sounded like a little kid" & \textbf{Professional and Mature:} Preference for the robot character to be professional and mature, so its feedback on LVR can be trustworthy. \\
\hline
"humorous videos about vision"; "more humor" & \textbf{Friendly and Enjoyable:} Preference for the robot character to be fun and amusing.\\
\hline
"AI interaction"; "evaluating the quality of my reading"; “personalize the robot’s responses, like saying my name during conversations”; "give feedback, and show specific interest in the user" & \textbf{Intelligent and Personalized:} Preference for the SAR to use artificial intelligence to perceive users' reading ability and provide personalized feedback accordingly \\
\hline
“Give more encouraging statements”; "Give information about new devices that can help patients see" & \textbf{Encouraging and Empathetic:} Preference for the SAR to inspire and motivate users with LV to be persistent with LVR.\\

\hline

“program it to have more consistency, ask more questions, give feedback, and show specific interest in the user” & \textbf{Long-Term:} Preference for the SAR to interact with users consistently over time and longitudinally with incremental advice.\\

\hline

\hline
\end{tabularx}

\label{tab:design}
\end{table*}

\textbf{Identified Design Guidelines: } Based on the responses collected from the semi-structured interview questions surveyed in the Phase 2 in-clinic demo (N=13), we followed the process of inductive coding~\cite{thomas2006general} to identify and summarize the most important user-informed design principles with corresponding proposed solutions in Table~\ref{tab:design}.

\section{Discussion}
\label{sec:discussion}

This work produced the following key insights.

\textit{Surveyed older adult users of magnifiers for LV stated that a SAR could provide helpful instructions for magnifier learning, and for socio-emotional support to reduce the commonly reported frustration they experience during magnifier learning.} As reported in Section~\ref{sec:results} - H1, in both phases combined (N=51), the majority felt frustrated while using a magnifier. This suggests the need to develop the SAR intervention to provide not only educational instructions but also socio-emotional support to help motivate users' perseverance with the magnifier.


\textit{Older adults with LV who had no prior interaction with a SAR system for LV perceived that a SAR would be useful for helping with vision loss, particularly for those with self-reported very poor vision.} As detailed in Section~\ref{sec:results} - H2, consistent with our hypotheses, the majority of the Phase 1 study participants (N=38) believed the SAR would be useful with managing vision loss, a good idea to facilitate LVR, trustworthy, easy to use, and enjoyable. In addition, participants who had the worst ratings of their vision were more likely to think that a SAR was a good idea to help with managing vision loss. Brought together, these results provide evidence and motivation for the development of novel SAR interventions to help users with visual impairment.

\textit{Older adults with LV who took part in an in-clinic demo interaction with our designed prototype SAR reported more positive opinions of the SAR's capabilities than those who only filled out mailed-in surveys without a SAR demo.} As reported in Section~\ref{sec:results} - H3, there were no significant differences between groups for perceived usefulness or trustworthiness, which indicates the interactions with the SAR system likely met expectations for those criteria. Moreover, our results revealed that participants who had the demo interaction were more likely to indicate the SAR was easy to use or enjoyable. Therefore, a SAR demo may help to improve potential LV users' perceptions of those desirable SAR attributes.


\textit{A set of design principles were identified to inform future development of SARs for this novel domain.} From the semi-structured interview responses obtained in the in-clinic SAR demos with LV participants, we found that participating older adults with LV preferred the SAR's character to be: 1) professional and mature; 2) friendly and enjoyable; 3) intelligent and personalized; 4) encouraging and empathetic; and 5) long-term. More details about the summarized design principles and participants' quotes can be found in Table~\ref{tab:design}.

\section{Conclusion}
\label{sec:conclusion}

This work employed a user-informed design process to validate the motivation and major design principles for developing a SAR for LVR interventions. Based on the quantitative and qualitative data collected in a two-phase survey study, we found significant evidence supporting the future development of SAR to address the socio-emotional needs of older adults with LV because such potential users reported that it would be useful, trustworthy, and enjoyable to use a SAR to augment LVR interventions. Future work may benefit from the resulting design principles derived from participant feedback to inform the development of personalized, autonomous SARs for older adults with LV in long-term in-home settings. \\
\textit{\textbf{ Acknowledgements.}} This work was supported by the NIH grant for "Development of a Behavioral Intervention with Socially Assistive Robots to Enhance Magnification Device Use for Reading" (5R21EY031126-02) and the University of Southern California.

\bibliographystyle{splncs04}
\bibliography{references}

\begin{thebibliography}{10}
\providecommand{\url}[1]{\texttt{#1}}
\providecommand{\urlprefix}{URL }
\providecommand{\doi}[1]{https://doi.org/#1}

\bibitem{qtrobot}
“qtrobot: Humanoid social robot for research and teaching.”.
  \url{https://luxai.com/humanoid-social-robot-for-research-and-teaching},
  accessed: 2010-09-30

\bibitem{abdi2018scoping}
Abdi, J., Al-Hindawi, A., Ng, T., Vizcaychipi, M.P.: Scoping review on the use
  of socially assistive robot technology in elderly care. BMJ open
  \textbf{8}(2),  e018815 (2018)

\bibitem{beuscher2017socially}
Beuscher, L.M., Fan, J., Sarkar, N., Dietrich, M.S., Newhouse, P.A., Miller,
  K.F., Mion, L.C.: Socially assistive robots: measuring older adults'
  perceptions. Journal of gerontological nursing  \textbf{43}(12),  35--43
  (2017)

\bibitem{binns2012effective}
Binns, A.M., Bunce, C., Dickinson, C., Harper, R., Tudor-Edwards, R.,
  Woodhouse, M., Linck, P., Suttie, A., Jackson, J., Lindsay, J., et~al.: How
  effective is low vision service provision? a systematic review. Survey of
  ophthalmology  \textbf{57}(1),  34--65 (2012)

\bibitem{bittner2018feasibility}
Bittner, A.K., Yoshinaga, P., Bowers, A., Shepherd, J.D., Succar, T., Ross,
  N.C.: Feasibility of telerehabilitation for low vision: satisfaction ratings
  by providers and patients. Optometry and Vision Science  \textbf{95}(9),
  865--872 (2018)

\bibitem{chiang2011global}
Chiang, P.P.C., O’Connor, P.M., Le~Mesurier, R.T., Keeffe, J.E.: A global
  survey of low vision service provision. Ophthalmic epidemiology
  \textbf{18}(3),  109--121 (2011)

\bibitem{chodorow2013mongodb}
Chodorow, K.: MongoDB: the definitive guide: powerful and scalable data
  storage. " O'Reilly Media, Inc." (2013)

\bibitem{Deng2019EmbodimentIS}
Deng, E., Mutlu, B., Mataric, M.: Embodiment in socially interactive robots.
  ArXiv  \textbf{abs/1912.00312} (2019)

\bibitem{dougherty2011abandonment}
Dougherty, B.E., Kehler, K.B., Jamara, R., Patterson, N., Valenti, D.,
  Vera-Diaz, F.A.: Abandonment of low vision devices in an outpatient
  population. Optometry and vision science: official publication of the
  American Academy of Optometry  \textbf{88}(11), ~1283 (2011)

\bibitem{gadde2011toward}
Gadde, P., Kharrazi, H., Patel, H., MacDorman, K.F.: Toward monitoring and
  increasing exercise adherence in older adults by robotic intervention: a
  proof of concept study. Journal of Robotics  \textbf{2011} (2011)

\bibitem{goldstein2012baseline}
Goldstein, J.E., Massof, R.W., Deremeik, J.T., Braudway, S., Jackson, M.L.,
  Kehler, K.B., Primo, S.A., Sunness, J.S., Group, L.V.R.N.S., et~al.: Baseline
  traits of low vision patients served by private outpatient clinical centers
  in the united states. Archives of Ophthalmology  \textbf{130}(8),  1028--1037
  (2012)

\bibitem{liu2013occupational}
Liu, C.J., Brost, M.A., Horton, V.E., Kenyon, S.B., Mears, K.E.: Occupational
  therapy interventions to improve performance of daily activities at home for
  older adults with low vision: A systematic review. American Journal of
  Occupational Therapy  \textbf{67}(3),  279--287 (2013)

\bibitem{mathew2014overview}
Mathew, S., Varia, J.: Overview of amazon web services. Amazon Whitepapers
  (2014)

\bibitem{pollard2003barriers}
Pollard, T.L., Simpson, J.A., Lamoureux, E.L., Keeffe, J.E.: Barriers to
  accessing low vision services. Ophthalmic and Physiological Optics
  \textbf{23}(4),  321--327 (2003)

\bibitem{quigley2009ros}
Quigley, M., Conley, K., Gerkey, B., Faust, J., Foote, T., Leibs, J., Wheeler,
  R., Ng, A.Y., et~al.: Ros: an open-source robot operating system. In: ICRA
  workshop on open source software. vol.~3, p.~5. Kobe, Japan (2009)

\bibitem{rantanen2017home}
Rantanen, P., Parkkari, T., Leikola, S., Airaksinen, M., Lyles, A.: An in-home
  advanced robotic system to manage elderly home-care patients’ medications:
  A pilot safety and usability study. Clinical therapeutics  \textbf{39}(5),
  1054--1061 (2017)

\bibitem{rubin2013measuring}
Rubin, G.S.: Measuring reading performance. Vision research  \textbf{90},
  43--51 (2013)

\bibitem{short2020sprite}
Short, E.S., Short, D., Fu, Y., Matari{\'c}, M.: Sprite: Stewart platform robot
  for interactive tabletop engagement. arXiv preprint arXiv:2011.05786  (2020)

\bibitem{stelmack2017outcomes}
Stelmack, J.A., Tang, X.C., Wei, Y., Wilcox, D.T., Morand, T., Brahm, K.,
  Sayers, S., Massof, R.W., Group, L.I.S., et~al.: Outcomes of the veterans
  affairs low vision intervention trial ii (lovit ii): a randomized clinical
  trial. JAMA ophthalmology  \textbf{135}(2),  96--104 (2017)

\bibitem{thomas2006general}
Thomas, D.R.: A general inductive approach for analyzing qualitative evaluation
  data. American journal of evaluation  \textbf{27}(2),  237--246 (2006)

\end{thebibliography}





\end{document}